\documentclass{article}

\usepackage[main, final]{neurips_2026}

\usepackage[utf8]{inputenc} 
\usepackage[T1]{fontenc}    
\usepackage{hyperref}       
\usepackage{url}            
\usepackage{booktabs}       
\usepackage{graphicx}       
\usepackage{amsfonts}       
\usepackage{amsmath}        
\usepackage{nicefrac}       
\usepackage{microtype}      
\usepackage{xcolor}         

\usepackage{pifont}
\newcommand{\cmark}{\ding{51}}  
\newcommand{\xmark}{\ding{55}}  
\usepackage{makecell}

\title{ST-Bench: A Spatial-Temporal Benchmark for Multi-Agent System Generation on Scientific Research Tasks}

\author{
  Qi Cheng \\
  Rutgers University\\
  New Brunswick, NJ\\
  \texttt{q.cheng@rutgers.edu} \\
  \And
  Rongchao Dong\\
  Rutgers University\\
  New Brunswick, NJ \\
  \texttt{rc.dong@rutgers.edu} \\
  \AND
  Shengyu Chen \\
  University of Pittsburgh \\
  Pittsburgh, PA \\
  \texttt{shc160@pitt.edu} \\
  \And
  Licheng Liu \\
  University of Minnesota \\
  Minneapolis, MN \\
  \texttt{lichengl@umn.edu} \\
  \And
  Dan Lu \\
  Oak Ridge National Laboratory\\
  Oak Ridge, TN \\
  \texttt{lud1@ornl.gov} \\
  \And
  Zhengzhang Chen \\
  NEC Labs America\\
  Princeton, NJ \\
  \texttt{zchen@nec-labs.com} \\
  \And
  Wei Cheng \\
  NEC Labs America\\
  Princeton, NJ \\
  \texttt{weicheng@nec-labs.com} \\
  \And
  Yiqun Xie \\
  University of Maryland\\
  College Park, MD \\
  \texttt{xie@umd.edu} \\
  \And
  Haifeng Chen \\
  NEC Labs America\\
  Princeton, NJ \\
  \texttt{haifeng@nec-labs.com} \\
  \And
  Xiaowei Jia \\
  Rutgers University\\
  New Brunswick, NJ\\
  \texttt{xj159@cs.rutgers.edu} \\
  \And
  Haoyu Wang \\
  NEC Labs America\\
  Princeton, NJ \\
  \texttt{haoyu@nec-labs.com} \\
}

\begin{document}

\maketitle

\begin{abstract}
The rapid progress of LLM-based multi-agent systems (MAS) has shown that they largely outperform single agents on coding, math, and QA tasks, where executable tests provide a binary success signal. Whether this advantage transfers to real scientific data analysis remains untested. We introduce ST-Bench, a benchmark designed to answer two questions: whether MAS outperform single agents on complex scientific data analysis tasks, and if so, by how much and at what additional cost. ST-Bench contains 100 data science tasks adapted from published Earth science studies across hydrology, agriculture, and wetland methane research, expanded into 2,067 queries grounded in additional published studies and validated by domain experts. Using ST-Bench, we evaluate five recent MAS generation methods under two training protocols, against single-agent baselines on the same GPT-5 backbone. Nine of the ten MAS configurations exceed the cheapest single-agent baseline, with the strongest reaching nearly three times its composite score. This gain is primarily attributable to coverage: trained workflows produce realistic numerical metrics on a larger fraction of queries, while the quality of those metrics, conditional on producing realistic output, is comparable to that of the single-agent baseline. The strongest configuration requires approximately four times the single-agent inference time, whereas a more economical workflow captures the majority of the benefit at less than twice the cost. MAS specialization confers measurable benefit on scientific data analysis, but the benefit is conditional rather than universal.
\end{abstract}

\section{Introduction}
\label{sec:intro}

To amplify the capabilities of LLMs, a substantial series of recent works have orchestrated LLMs into multi-agent systems (MAS), in which a set of specialized agents coordinate through designed workflows to solve tasks more reliably than single agents. Early MAS frameworks established the design space through hand-crafted role decompositions and communication protocols~\citep{camel,autogen,metagpt,multipersona,selfrefine}. More recent work treats the workflow itself as an object to be optimized, which automatically search for strong workflows over a target task distribution ~\citep{aflow, adas, gptswarm, metaagent, w4s, flowreasoner, maas}. Across coding benchmarks ~\citep{humaneval, mbpp}, math benchmarks \citep{gsm8k, math}, and question answering benchmarks~\citep{hotpotqa, drop}, these methods report substantial improvements over single-agent baselines. Building on these results, it is natural to ask whether MAS approaches can also advance automated scientific research, where tasks are often more complex, data-centric, and open-ended. 

\begin{figure}[t]
  \centering
  \includegraphics[width=\textwidth]{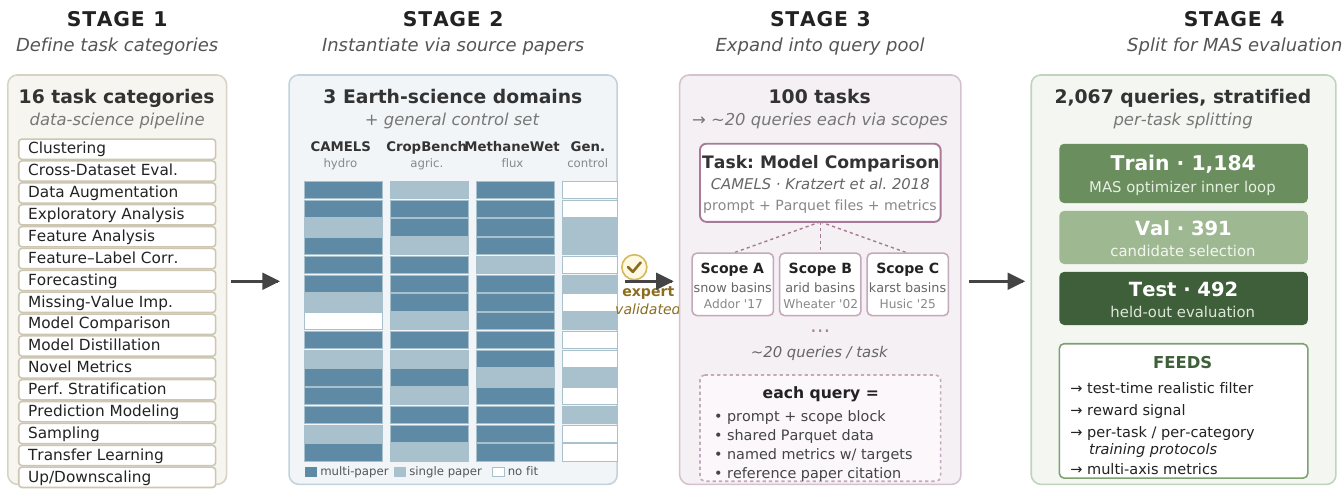}
  \vspace{-20pt}
  \caption{\textbf{Construction of ST-Bench.} We first define 16 task categories spanning the data-science pipeline (Stage 1), then instantiate each (category, domain) cell with one or more peer-reviewed source papers (Stage 2). Every resulting task is expanded to ${\sim}20$ paper-grounded queries via dataset scopes (Stage 3). The 2{,}067 queries are stratified into per-task train/val/test splits that feed both the MAS optimizer and our evaluation framework (Stage 4).}
  \label{fig:construction}
  \vspace{-15pt}
\end{figure}

In this work, we argue that no existing MAS benchmarks can be used to answer this research question. Most benchmarks used to validate 
MAS methods 
share a structural property that scientific research often lacks: a binary oracle. HumanEval and MBPP grade through executable unit tests, and GSM8K and MATH grade through a single numerical answer. Each candidate workflow can therefore be assigned a clean binary score, which is precisely the 
feedback signal that MAS inner-loop optimizers were designed to use. Real scientific data analysis tasks are much more diverse and often 
do not offer such signal. A clustering task can produce many valid clusterings, a forecasting task can be partially correct along multiple numerical axes, and a gap filling task can produce a usable answer that does not match any single reference exactly. 
In these settings, success is better measured through continuous or partial-credit metrics rather than binary correctness. It therefore remains an open empirical question whether workflow specialization, which has been shown to be effective under binary-oracle benchmarks, still provides a measurable advantage for diverse scientific tasks evaluated with more nuanced scientific metrics.

Although several benchmarks for scientific tasks already exist, they do not support such a test. 
Scientific agent benchmarks ~\citep{ScienceAgentBench, blade, discoverybench} provide carefully curated scientific tasks, but were designed to evaluate a single agent on a single attempt per task. They therefore lack the within-task pool of samples that MAS generation methods require: an inner-loop optimizer cannot iterate on a single query, and a held-out single query cannot test whether an optimized workflow generalizes. Spatial-temporal scientific benchmarks ~\citep{PDEBench, CAMELS} provide the real heterogeneous data that scientific tasks involve, but they were designed to compare fixed model classes or forecasting skills, not LLM-generated analysis workflows, and lack the prompt and metric that lets a generated MAS read raw files, choose preprocessing steps, run analysis code, and report named scientific metrics. 
Hence, the
comparative question we wish to answer cannot be answered by any available benchmark. A comparison between available benchmarks can be found in Table~\ref{tab:benchmark-comparison} and extended related work is provided in Appendix~\ref{app:related-work}.. 

To this end, we present ST-Bench, a benchmark and evaluation framework designed to enable rigorous comparison both between MAS generation systems and against single-agent baselines on complex scientific research tasks. The construction of ST-Bench follows four design principles. (1) \textit{Tasks grounded in real published scientific work:} We adapt 100 tasks from peer-reviewed Earth science publications spanning three science domains. Each task has processed data files, a natural language prompt, and a list of metrics providing what spatial-temporal data benchmarks lack. (2) \textit{Query pools per task:} We expand each task into approximately twenty queries by attaching dataset scopes drawn from additional peer-reviewed publications. This within-task pool supplies the data samples that MAS generation methods require to optimize, while the test queries let them do evaluations, neither of which is possible on current agentic benchmarks. (3) \textit{Per-task reward signals:} Multi-agent generation methods rely on optimizers that select between candidate workflows by using a score per candidate. Existing task specific benchmarks often provide this signal through unit tests, but existing scientific benchmarks do not. We therefore equip every task in ST-Bench with a continuous reward signal, which is a realistic filter that excludes defensive defaults paired with a reference-aware verifier returns a continuous scorer. To our knowledge, ST-Bench is the first scientific benchmark whose per-task signals are fine-grained enough to drive MAS workflow optimization on complex science problems rather than only grading a final attempt. (4) \textit{A multi-axis evaluation framework:} ST-Bench reports performance along two orthogonal sets of axes. The first is task structure: the 100 tasks span 16 categories covering the standard data science pipeline, three difficulty tiers, and four analytical tag sets recording the capability dimension, pipeline stage, cognitive skill, and MAS-specific challenge that each task most stresses. The second is system property where a per-task win rate against the single-agent baseline, cost ratios in both inference time and token consumption, an efficiency score expressing accuracy gain per unit of additional cost, a generalizability measure that compares workflows trained at per-task and per-category granularity, and additional axes for scalability across LLM backbones and internal workflow parallelism. Together, these tasks and metric axes let ST-Bench produce a comprehensive diagnostic of where each MAS succeeds, where it fails, and along which dimensions its specialization is worth its cost. The construction pipeline are shown in Figure ~\ref{fig:construction}.
 
Using ST-Bench, we evaluate five recent MAS generation methods against single-agent baselines. Comparison against the single agent addresses the inter-system half of our research goal. Comparison among the five methods, and between each method's per-task and per-category variants, addresses the intra-system half. We also report cost along two dimensions, wall-clock inference time and token consumption where token accounting is available. Our contribution is therefore not a claim that current MAS methods solve scientific data analysis, but a benchmark and evaluation framework together with empirical findings that establish, both within MAS generation methods and against single-agent baselines, when these systems improve, by how much, and at what cost. We position ST-Bench as a measurement tool for the open question of when MAS is worth its added compute, and as a way to provide insights for future MAS designs that can compute over real heterogeneous data, score real scientific metrics, and manage inference cost effectively.

\begin{table*}[t]
\centering
\footnotesize
\caption{Comparison of ST-Bench to agent benchmarks. \textbf{MAS Opt.} indicates whether the benchmark provides within-task query pools for MAS generation systems. \textbf{Reward Signal} distinguishes between a binary oracle and a continuous reward. \textbf{Multi-axis Eval.} indicates evaluation along orthogonal axes beyond accuracy.}
\label{tab:benchmark-comparison}
\resizebox{\textwidth}{!}{
\begin{tabular}{lccccc}
\toprule
\textbf{Benchmark} & \makecell{\textbf{Task} \\ \textbf{Sources}} & \makecell{\textbf{Code} \\ \textbf{Gen.}} & \makecell{\textbf{MAS} \\ \textbf{Opt.}} & \makecell{\textbf{Reward} \\ \textbf{Signal} } & \makecell{\textbf{Multi-axis} \\ \textbf{Eval.}} \\
\midrule
SWE-Bench~\citep{jimenez2023swe}            & GitHub          & File-level   & \xmark & Binary     & \xmark \\
AgentBench~\citep{liu2023agentbench}             & Mixed           & Mixed        & \xmark & Binary     & \xmark \\
TaskBench~\citep{shen2024taskbench}              & Synthetic       & No code      & \xmark & Binary     & \xmark \\
\midrule
MLAgentBench~\citep{huang2023mlagentbench}       & Kaggle          & File-level   & \xmark & Continuous & \xmark \\
BLADE~\citep{blade}                              & 31 Publications & Function     & \xmark & Binary     & \xmark \\
DiscoveryBench~\citep{discoverybench}            & 27 Publications & Code Gen     & \xmark & Binary     & \xmark \\
ScienceAgentBench~\citep{ScienceAgentBench}      & 44 Publications & File-level   & \xmark & Mixed      & Partial \\
\midrule
\textbf{ST-Bench (Ours)}                         & \textbf{Publications} & \textbf{File-level} & \cmark & \textbf{Continuous} & \cmark \\
\bottomrule
\end{tabular}
}
\vspace{-15pt}
\end{table*}

\section{Dataset}
\label{sec:dataset}

In this section, we introduce ST-Bench, which aims to evaluate MAS generation systems and single-agent baselines on the real research tasks that involve large-scale spatio-temporal data. We therefore anchor every task in ST-Bench to a peer-reviewed Earth science publication. That publication defines the scientific question, identifies the data on which it is studied, and defines the metric used to evaluate performance. Each task is then turned into a MAS readable instance that includes a natural language prompt, a directory of preprocessed uni-format files, and a list of named scientific metrics. 
Hence, ST-Bench moves beyond synthetic tests of MAS coordination on toy problems and instead evaluates whether a generated MAS can solve scientific problems to the standards established by peer-reviewed research.

\subsection{Problem Formulation}
\label{sec:dataset-formulation}

Given a natural language prompt and a directory of raw data files, a system under evaluation must read and process the data, decompose the task, run the analysis the task requires, and output related metric values. We resolve the task at the level of a single query rather than a free-form code submission, so that the same machinery used to grade a final attempt can also drive an MAS optimizer. Here we introduce the four components of each query, each chosen to support both single-agent inference and MAS workflow generation.

\paragraph{Task Prompt.} 
A task prompt that over-specifies the algorithm short-circuits the workflow generation problem we aim to study, while one that under-specifies the goal leaves no measurable target. We therefore phrase each prompt as a concise scientific objective and avoid prescribing a particular algorithmic approach. Open-ended prompts force a generated MAS to actually decompose the task and select tools, rather than retrieve a memorized solution from a coding benchmark. They also leave the metric space large enough that the optimizer can find improvements rather than running into a single optimum on the first iteration. Each prompt also contains a block that specifies the subset of the dataset to be used. 

\paragraph{Dataset Files.} Raw scientific datasets ship in many file formats, and a benchmark that exposed every published format directly would force tested MAS systems to spend more effort on format parsing than on scientific reasoning. We therefore convert every dataset we used and their subsets to a uniform Parquet representation before staging it into the working directory, so that file-format handling is not the variable under test.

\paragraph{Defined Metrics.}The MAS optimizer's inner loop requires a continuous, partial-credit fitness signal per task; a binary oracle does not transfer to scientific data analysis. We therefore equip each task with a list of named numeric metrics, each paired with an inequality target such as ``$> 0.3$'' (when higher values are better) or ``lower is better'' (when no fixed threshold is specified). Inequality targets provide two advantages over gold answers (i.e., single ground-truth answers). 
First, they accommodate the open-endedness of valid scientific solutions: many different clusterings can clear a Silhouette $> 0.3$ bar, so the candidate space the MAS generation method optimizes over is genuinely large rather than a single retrieval. Second, the numeric value the system reports, not just whether it cleared the target, is a continuous quantity, which is exactly what an inner-loop fitness function needs to differentiate between candidate workflows that all happen to pass the target. The training-time verifier (\S\ref{sec:eval-verifier}) consumes this continuous quantity directly.

\paragraph{Reference.}
We attach to each task a citation to the publication it was adapted from. Note that this field 
is stored only for expert validation and for analysis of benchmark coverage, and it is not exposed to MAS generation systems and single agent baselines.

\subsection{Dataset Construction}
\label{sec:dataset-construction}
 
ST-Bench was built top-down rather than by accumulating tasks from a single source. We first fixed task categories from data science capabilities, then filled each category with papers from three real Earth science domains, and expanded each resulting task into a pool of paper-grounded queries. The end product is 100 tasks expanded into 2{,}067 queries. Figure~\ref{fig:dataset-composition} summarizes the resulting task and query distribution.
 
 
\paragraph{Task Categories.} Before any source paper or dataset was selected, we defined 16 task categories spanning the standard data science pipeline: Clustering, Cross-Dataset Evaluation, Data Augmentation, Exploratory Data Analysis, Feature Analysis, Feature and Label Correlation Checking, Forecasting Modeling, Missing Value Imputation, Model Comparison, Model Distillation, Novel Evaluation Metrics, Performance Stratification, Prediction Modeling, Sampling Strategies, Transfer Learning, and Up/Downscaling. The categories cover data preparation, exploration, modeling, and evaluation, so every internal role a generated MAS might specialize in is exercised on multiple tasks. They are also concrete enough that two source papers in the same category produce comparable problems, which lets per-category training (\S\ref{sec:eval-protocols}) operate on a coherent training pool.
 
\paragraph{Science Domains.}
With the categories fixed, we chose three Earth science domains to populate them.
We instantiate the 16 categories across three Earth-science domains: hydrology using CAMELS ~\citep{CAMELS}, agriculture using two crop yield datasets ~\citep{cropbench1, cropbench2}, and wetland methane using X-MethaneWet/FLUXNET-CH4 ~\citep{c7.5-7.6.4-9.5.3}, plus 11 general-control tasks. These domains provide heterogeneous spatial-temporal data and cover forecasting, clustering, transfer, model comparison, gap filling, and stratified evaluation. Full dataset descriptions and source details are provided in Appendix~\ref{app:domains}.

\begin{table}[t]
\centering
\small
\caption{A representative task in ST-Bench: C3.4 (Model Comparison, hydrology).}
\begin{tabular}{p{0.18\textwidth}p{0.78\textwidth}}
\toprule
Field & Value \\
\midrule
Task ID         & C3.4 \\
Difficulty      & Easy \\
Category        & Model Comparison \\
Domain          & Hydrology (CAMELS) \\
Source          & Kratzert et al.\ (2018), HESS~\citep{example_query}\\
\midrule
Prompt          & You are given daily meteorological forcing and observed streamflow for 10 CAMELS basins spanning diverse hydrological regimes. Train an LSTM, a random forest, and a linear regression model independently for each basin, evaluate them using NSE, KGE, and the flow duration curve, and analyze residual patterns where the three models disagree. Report the median NSE across basins and the cross-basin consistency of model rankings. The data files can be found in \{Scope file path\}\\
\midrule
Metrics         & Median NSE across basins; model ranking consistency across basins. \\
\bottomrule
\label{tab:example-task}
\end{tabular}
\vspace{-20pt}
\end{table}

\paragraph{Tasks.}
For each (category, domain) pair, we sourced one or more peer-reviewed publications that operationalized the category's analytical skill on that domain's data. Some cells admitted multiple natural instances and yielded several tasks; a few cells had no natural fit and are not represented. For each chosen source paper we wrote a concise prompt, reformulated quantitative success criteria as named metrics with inequality targets, and prepared the staged Parquet files needed to compute those metrics. The eventual count of tasks per category ranges from 4 (Model Distillation) to 11 (Feature and Label Correlation Checking). Table~\ref{tab:example-task} shows one task end-to-end.

\paragraph{Queries}
A single source paper gives one analysis on one canonical dataset slice. That is sufficient for grading a single-agent attempt, but it leaves an MAS generation method that operate on task-level with no examples to train on. We therefore expand each task into approximately twenty queries by sourcing additional peer-reviewed publications that defined named subsets of the underlying dataset, and attaching one such subset to each query as its scope. A scope record specifies a named geographic, temporal, or sample-size slice together with a short description, a numeric size, and a citation to the publication that defined that slice. The processed data files are identical across queries within a task. The scope changes which subset of those files the system is asked to operate on. Scope variation between train and test queries forces the MAS optimizer to seek workflows that generalize across data samples rather than ones tuned to a specific subset.
 
\paragraph{Splits.}
The 2{,}067 queries are partitioned into 1{,}184 train, 391 validation, and 492 test queries, stratified by task so that every task has at least three test queries and a non-trivial train and validation pool. 
\begingroup
\setlength{\abovedisplayskip}{3pt plus 1pt minus 1pt}
\setlength{\belowdisplayskip}{3pt plus 1pt minus 1pt}
\setlength{\abovedisplayshortskip}{2pt plus 1pt minus 1pt}
\setlength{\belowdisplayshortskip}{2pt plus 1pt minus 1pt}

\section{Evaluation Framework}
\label{sec:eval}

We evaluate whether MAS generation systems outperform single-agent baselines on real scientific data-driven tasks, and whether any gain justifies the added cost. This requires three components: a test-time grader that rewards real numerical computation rather than defensive defaults, a continuous training-time reward signal for MAS optimizers, and cost-aware metrics that distinguish better metric quality from broader query coverage. Additional details are provided in Appendix~\ref{app:eval-framework}.

\subsection{Training Protocols}
\label{sec:eval-protocols}

We evaluate every MAS method under two training protocols. In the \emph{per-task} protocol, a method optimizes separately on each of the 100 tasks using only that task's train and validation queries, producing 100 task-specific MAS instances for task-level MAS generation system. In the \emph{per-category} protocol, a method optimizes on the union of train and validation queries from all tasks in a task category, producing 16 category-level MAS instances for task-level MAS generation system. Both protocols are evaluated on the same 492 held-out test queries (\S\ref{sec:dataset-construction}). 

\subsection{Evaluation Metrics}
\label{sec:eval-axes}

Our evaluation reports one absolute performance score and several comparative metrics relative to the single-agent baseline. Let $M$ denote a generated MAS configuration, $SA$ the single-agent baseline, $T$ the set of tasks, and $\mathrm{score}_{X,t}$ the per-task score of system $X$ on task $t$.

\paragraph{Absolute performance score.}
For each task, we compute a success rate over realistic outputs and a pass rate over named scientific metrics. A query is treated as realistic only if it reports parsable finite metrics, includes at least one strictly positive numeric value, and does not use a $-1$ sentinel. The per-task score is
{\small
\[
\mathrm{score}_t \;=\; \mathrm{success\_rate}_t \,\cdot\, \overline{\mathrm{pass\_rate}}_t,
\]
}
where $\overline{\mathrm{pass\_rate}}_t$ is the mean metric pass rate over realistic test queries, set to zero when no realistic output is produced. The system-level score is the sum over all 100 tasks, with maximum 100.

\paragraph{Win rate and mean score gain.}
We measure how often and by how much a MAS configuration improves over the single-agent baseline:
{\small
\[
\mathrm{win\_rate}(M) \;=\; \frac{1}{|T|}\sum_{t \in T} \mathbb{1}\bigl[\mathrm{score}_{M,t} > \mathrm{score}_{SA,t}\bigr],
\quad
\Delta_{\mathrm{perf}}(M) \;=\; \frac{1}{|T|}\sum_{t \in T} \bigl(\mathrm{score}_{M,t} - \mathrm{score}_{SA,t}\bigr).
\]
}

\paragraph{Cost ratio.}
We report inference cost relative to the single-agent baseline using wall-clock time and, where available, token consumption:
{\small
\[
\rho_{\mathrm{time}}(M) = \frac{\bar{c}^{\mathrm{time}}_{M}}{\bar{c}^{\mathrm{time}}_{SA}}, \qquad
\rho_{\mathrm{token}}(M) = \frac{\bar{c}^{\mathrm{token}}_{M}}{\bar{c}^{\mathrm{token}}_{SA}},
\]
}
where $\bar{c}^{\cdot}_X$ is the mean per-query cost of system $X$.

\paragraph{Efficiency.}
We summarize cost-adjusted improvement as mean score gain per unit of additional inference time:
{\small
\[
\eta(M) \;=\; \frac{\Delta_{\mathrm{perf}}(M)}{\max\bigl(\rho_{\mathrm{time}}(M) - 1,\, 0.1\bigr)}.
\]
}
The denominator is clipped at $0.1$ to avoid instability when the cost ratio is close to one. The same definition applies to token cost when token accounting is available.

\paragraph{Generalizability.}
To compare narrow specialization with broader transfer, we measure the mean per-task minus per-category score gap:
{\small
\[
\mathrm{gen}(M) \;=\; \frac{1}{|T|}\sum_{t \in T} \bigl(\mathrm{score}_{M^{\mathrm{task}}, t} - \mathrm{score}_{M^{\mathrm{cat}}, t}\bigr),
\]
}
where $M^{\mathrm{task}}$ and $M^{\mathrm{cat}}$ are the per-task and per-category variants of the same method. Positive values indicate that narrow task-specific optimization helps; near-zero or negative values indicate that category-level workflows transfer as well or better.

Together, these metrics distinguish absolute performance, breadth of improvement, cost, cost-adjusted efficiency, and training-granularity generalization. Because the 100 tasks span heterogeneous scientific domains and objectives, we also annotate each task with difficulty and four tag axes---capability dimension, pipeline stage, cognitive skill, and MAS challenge---for post-hoc diagnostic cuts. Full label definitions, counts, and examples are provided in Appendix~\ref{app:task-tags}.

\subsection{Training-time Reward Function}
\label{sec:eval-verifier}

MAS optimizers require a scalar reward per candidate workflow, but scientific data analysis lacks the binary oracle used in coding benchmarks. We therefore use a continuous reference-aware verifier during training. For each reported metric, the reward is
{\small
\[
s \;=\; 0.10 \cdot s_{\mathrm{sanity}} \;+\; 0.40 \cdot s_{\mathrm{target}} \;+\; 0.50 \cdot s_{\mathrm{reference}},
\]
}
where $s_{\mathrm{sanity}}$ checks finite numeric validity, $s_{\mathrm{target}}$ checks whether the metric satisfies its inequality target, and $s_{\mathrm{reference}}$ measures closeness to a cached single-agent reference value, clipped to $[0,1]$ and polarity-adjusted for lower-is-better metrics. To avoid rewarding defensive defaults, zero outputs against non-zero references receive zero reward. When the single-agent reference is unavailable, we drop $s_{\mathrm{reference}}$ and renormalize the remaining terms. The reward replaces each method's usual training-time scoring hook, while the prompt shown to the candidate workflow remains unchanged.

\endgroup
\section{Experiments and Results}
\label{sec:experiments}

\subsection{Experimental Setup}
\label{sec:exp-setup}

\paragraph{Backbone and baselines.}
All MAS generation methods and the primary single-agent baseline use GPT-5 with identical max-token and temperature settings, so any difference between systems is attributable to the workflow rather than the model. We additionally report baselines on GPT-5.2 and GPT-4o.

\paragraph{Methods and budgets.}
We evaluate five MAS generation methods selected to span the two major categories in the literature and a range of optimization mechanisms within them. Task-level methods optimize one workflow per problem and reuse it across queries within that problem; we include AFlow~\citep{aflow}, GPTSwarm~\citep{gptswarm}, and MetaAgent~\citep{metaagent}, which apply Monte Carlo tree search over operator graphs, REINFORCE-style edge-realization, and one-shot finite-state-machine generation, respectively. Query-level methods produce a fresh workflow per query through a meta-agent or an in-context controller; we include W4S~\citep{w4s} as a low-compute reinforcement-learned meta-agent and AutoAgents as a prompting-only Explorer baseline. Table~\ref{tab:budgets} reports the per-task settings.


\paragraph{Test set.}
All scores are computed on the 492 held-out test queries with the absolute performance score and realistic filter. 

\subsection{Comparison with the Single-Agent Baseline}
\label{sec:exp-headline}

\begin{figure}[t]
\centering
\includegraphics[width=\linewidth]{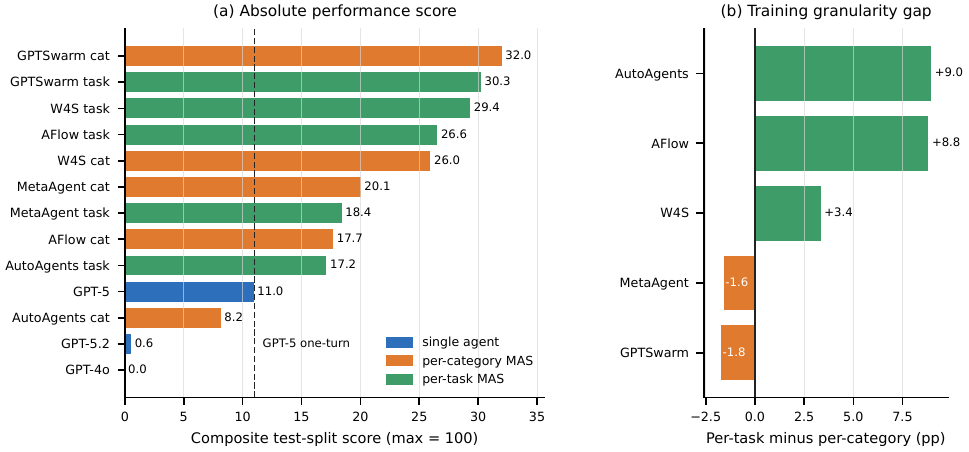}
\vspace{-15pt}
\caption{Absolute performance and training-granularity diagnostics. (a) Absolute performance score with the dashed line means the GPT-5 score. (b) Generalization gap, where positive values mean per-task training outperforms per-category training.}
\label{fig:headline-generalization}
\vspace{-15pt}
\end{figure}

Does multi-agent specialization improve over a strong single agent on real scientific data analysis? Figure~\ref{fig:headline-generalization}a reports the absolute performance score for all single-agent baselines and reward-function-trained MAS configurations on the 492 held-out test queries. The GPT-5 baseline scores 11.0 out of 100. Nine of the ten MAS configurations exceed it, with the strongest reaching 32.0 (GPTSwarm per-category), 30.3 (GPTSwarm per-task), 29.4 (W4S per-task), 26.6 (AFlow per-task), and 26.0 (W4S per-category). The single exception is AutoAgents per-category at 8.2, primarily because that configuration is capped at five test queries per category and leaves most queries unreached.

The number is only partially revealing of the performance. The MAS advantage on the absolute score is not that generated workflows are uniformly more accurate on every scientific metric. It is that some trained workflows produce realistic numerical metrics on a larger fraction of queries. Conditional on producing realistic output, the strongest MAS configurations clear a similar share of inequality targets as the single-agent baseline, typically between 50\% and 58\%. The MAS gain is therefore primarily a coverage gain: the workflow more often reaches the point at which scientific metrics can be computed at all, rather than producing higher-quality metrics once that point is reached.

\subsection{Cost-Performance Analysis}
\label{sec:exp-cost}

\begin{figure}[t]
\centering
\includegraphics[width=\linewidth]{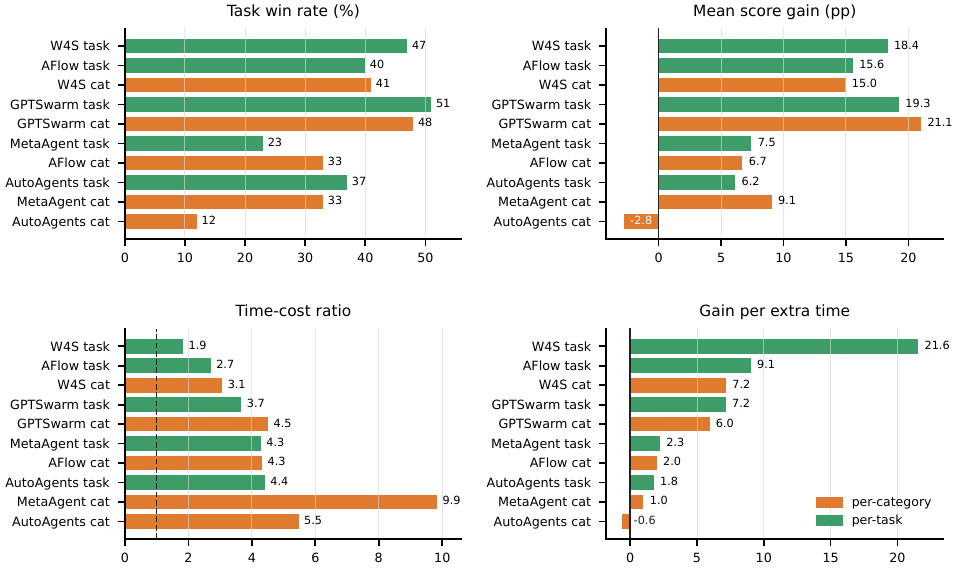}
\vspace{-15pt}
\caption{System-level metrics relative to GPT-5 one-turn. Task win rate measures how often the MAS improves a task's absolute score; mean score gain measures average absolute improvement; time-cost ratio measures wall-clock inference cost; efficiency is gain per unit of extra time.}
\label{fig:mas-axes}
\vspace{-0.6em}
\end{figure}

\begin{figure}[t]
\centering
\includegraphics[width=\linewidth]{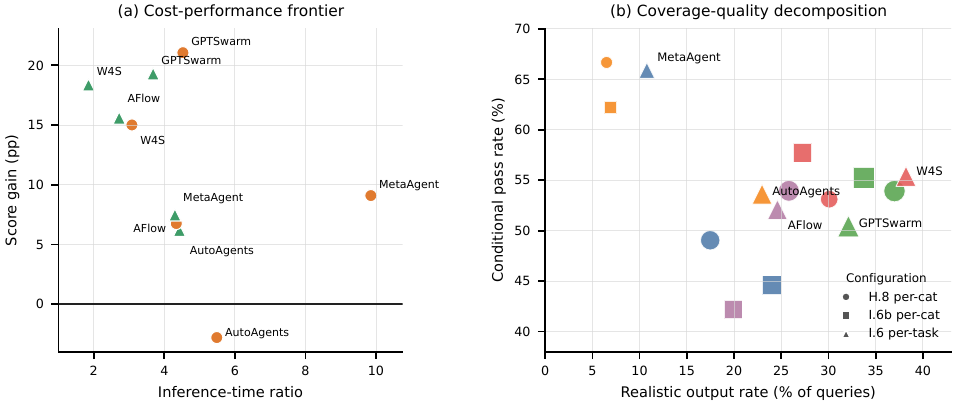}
\vspace{-15pt}
\caption{Cost and coverage-quality diagnostics. (a) Cost-performance frontier relative to GPT-5 one-turn, where the upper-left region is preferable; orange circles indicate per-category training and green triangles indicate per-task training. (b) Realistic output rate versus mean pass rate conditional on realistic output for each MAS configuration; marker size grows with the number of tasks that have at least one realistic output.}
\label{fig:cost-coverage}
\vspace{-0.6em}
\end{figure}

Does the inter-system gain justify the additional inference cost? The MAS configurations that beat the baseline pay between roughly $1.85$ and $4.5$ times the inference time, and the relative ordering of methods changes when this cost is taken into account. Figure~\ref{fig:mas-axes} reports the four primary system-level metrics relative to the reference, and Figure~\ref{fig:cost-coverage}a focuses on the cost-performance trade-off.

The strongest absolute-score configuration is GPTSwarm per-category, which gains $+21.1$ percentage points over the  baseline at roughly $4.5\times$ the inference time. GPTSwarm per-task achieves the highest task win rate ($51\%$) at a similar cost ratio. W4S per-task is the most efficient point on the frontier: it gains $+18.4$ percentage points at only $1.85\times$ the  inference time, yielding the highest gain per unit of additional compute. AFlow per-task is the next most efficient configuration. AutoAgents per-category is dominated; its cap on test queries leaves most queries unreached and produces a negative mean improvement despite nonzero quality on the queries it does attempt.

The cost view changes the interpretation of MAS success. If the deployment objective is maximum absolute score and inference cost is not constrained, GPTSwarm is preferred. If the objective is improvement per unit of additional compute, W4S per-task is preferred. No single MAS configuration dominates on both axes, and any production claim of MAS superiority therefore depends on which side of this trade-off the deployment lives on.

\subsection{Effect of Training Schema on Generalization}
\label{sec:exp-rq3}

Does training a single workflow on the union of sibling tasks generalize as well as training a separate workflow per task? The two protocols of Section~\ref{sec:eval-protocols} answer different generalization questions. Per-task training specializes a workflow narrowly while per-category training asks one workflow to transfer across sibling tasks in the same category. Figure~\ref{fig:headline-generalization}b reports the mean per-task minus per-category absolute-score gap, which is positive when narrow specialization wins and negative when category-level training transfers as well or better.

GPTSwarm and MetaAgent are near or below zero, meaning their category-level workflows generalize cleanly to held-out sibling tasks. AFlow and AutoAgents gain strongly from per-task specialization, suggesting that their workflows overfit or dilute when trained on a broader category pool. W4S sits between these regimes. The split is informative for MAS design: broader training pools help only when the induced workflow is abstract enough to transfer.

\subsection{Coverage-Quality Decomposition and Stratified diagnostics.}
\label{sec:coverage-quality}

To understand why MAS improves the absolute score, we decompose performance into realistic output rate and conditional pass rate. Figure~\ref{fig:cost-coverage}b shows that conditional pass rates are relatively clustered, typically around 50--58\%, while realistic output rates vary substantially across methods. Thus, the main MAS advantage is coverage: stronger workflows more often reach the point of producing realistic scientific metrics, rather than producing much better metrics once realistic outputs are obtained. Detailed method-level patterns and implications for future MAS design are provided in Appendix~\ref{app:coverage-quality}.

Additional stratified analyses show that MAS gains persist across all three scientific domains and across Easy, Medium, and Hard tasks, with especially large gains on data-preparation, robustness, decomposition, and parallel-search tasks. Full domain-, difficulty-, and tag-level results are provided in Appendix~\ref{app:exp-stratified}.

\section{Conclusion}
Scientific data analysis changes the rules under which generated multi-agent systems are evaluated. The bottleneck is not only reasoning depth or role decomposition; it is the absence of an oracle that can reward partial, real computation without rewarding trivial defaults. ST-Bench supplies that substrate through paper-grounded query pools, executable spatial-temporal datasets, realistic output filtering, and a reference-aware verifier. Our experiments show cautious upside: several generated workflows beat a one-turn GPT-5 baseline, but only under evaluation machinery that makes real data use visible. The benchmark's main value is therefore diagnostic. It lets future work ask which MAS designs genuinely compute, which merely format plausible metrics, and which forms of workflow specialization are worth their added cost.

\bibliographystyle{plainnat}
\bibliography{Qi_MAS,ref}

\appendix
\section*{Technical Appendix}
\section{Related Work}
\label{app:related-work}
\subsection{AI for Science}
Machine learning has been widely adopted across scientific domains~\cite{example_query,cropbench1,kgml3}. 
Although purely data-driven approaches can be accurate, they often suffer from limited interpretability, weak generalization, and outputs that violate real-world physical constraints~\cite{violating}.
Knowledge-Guided Machine Learning (KGML)~\cite{kgml1,kgml2,kgml4} addresses these issues by integrating domain-specific knowledge into model design, with representative directions including physics-informed neural networks~\cite{15.3.1,15.6.1}, physics-constrained generative models~\cite{15.1.2-15.4-15.5.1}, and foundation Earth models~\cite{c7.1.1-7.3.1-7.6.3,c7.3.2-7.6.1-7.7.1,kgml5}.
These advances center on building individual models for specific problems, leaving open how autonomous agents should orchestrate full data-driven scientific workflows that respect domain knowledge, e.g., feature engineering, model selection, evaluation, uncertainty quantification, and exploratory analysis. 
ST-Bench targets this gap by curating tasks that probe each stage of the data-driven discovery pipeline across three Earth-science domains.

\subsection{Domain Scientific Agents}
Beyond the general MAS frameworks discussed in Section~\ref{sec:intro}, a parallel line of work develops domain-specialized scientific agents. 
Single-agent systems such as Coscientist~\cite{singlesci1} and ChemCrow~\cite{singlesci2} pioneered tool-augmented reasoning for chemistry, while more recent multi-agent systems distribute responsibilities across specialized agents, including AI Co-Scientist~\cite{multiplesci1,multiplesci2} for parallel hypothesis evaluation, ProtAgents~\cite{multiplesci3} for protein discovery, GenoMAS~\cite{multiplesci4} for gene expression analysis, and The AI Scientist for end-to-end research automation~\cite{multiplesci5,multiplesci6}. 
These systems demonstrate the promise of role specialization, critical review, and multi-step planning for scientific tasks.
However, each is evaluated on narrow case studies tailored to its target domain, leaving the relative contribution of multi-agent coordination unmeasured in any standardized way. 
ST-Bench complements these efforts by providing a controlled evaluation suite that decouples MAS coordination from domain-specific tooling.

\section{Dataset}
\subsection{Science Domains}
\label{app:domains}
\textit{Hydrology.} CAMELS~\citep{CAMELS} dataset provides multi-decadal daily streamflow, meteorological forcing, and hundreds of static catchment attributes for over $\sim$700 minimally disturbed basins across the contiguous United States. We adapt 25 tasks reflecting core challenges studied in ML hydrology, including hydrological signature–based clustering, streamflow forecasting under climate and regional shift, signature regression, cross-basin transfer learning, and uncertainty stratification by basin elevation and humidity. CAMELS is known for strong heterogeneity across basins, sensitivity to nonstationarity, and difficulty of generalization to ungauged regions. These tasks require integrating static and dynamic data, handling long temporal dependencies, and constructing robust end-to-end analysis pipelines, making them a realistic test of whether MAS improves execution reliability beyond single-agent workflows.

\textit{Agriculture.} We compile a domain we refer to as CropBench from two publicly available sources: a county-level USA Corn Belt yield dataset associated with the CNN-RNN crop yield framework of \citet{cropbench1}, and the multi-region sub-national yield benchmark assembled by AgML CY-Bench~\citep{cropbench2}. Together they provide yield, management practice, soil composition, and weather records at the county-by-year level across multiple growing regions. We adapt 24 tasks, including yield prediction under management-regime shifts, weather-conditional yield clustering, model comparison across crop types, and stratified performance reporting by climate zone. CropBench tasks require disentangling confounding between management practices and yield outcomes, a recurring failure mode in naive tabular ML.
 
\textit{Wetland methane.} X-MethaneWet~\citep{c7.5-7.6.4-9.5.3} is a multi-site CH$_4$-flux compilation drawn from FLUXNET-CH4, with eddy-covariance flux measurements and environmental drivers across northern wetland sites. We adapt 40 tasks, including FLUXNET QC-flag agreement after data cleaning, gap-filling under instrument failure, regional versus pan-Arctic transfer, and site-stratified evaluation. The methane domain is uniquely demanding: fluxes are small relative to instrument noise, quality-control flags are binary expert annotations rather than physical measurements, and the underlying biogeochemistry has strong site-specific behavior that does not transfer through a single global model.
 
\textit{General control set.} We additionally include 11 non-domain tasks that operate on synthetic or general tabular data, such as clustering on isotropic Gaussian mixtures or feature-correlation on independent normals. We do not count General as a fourth domain. It serves as a negative control: a system that scores high on General but low on the three Earth science domains is failing the spatial-temporal task rather than the data science task.
\begin{figure}[t]
\centering
\includegraphics[width=\linewidth]{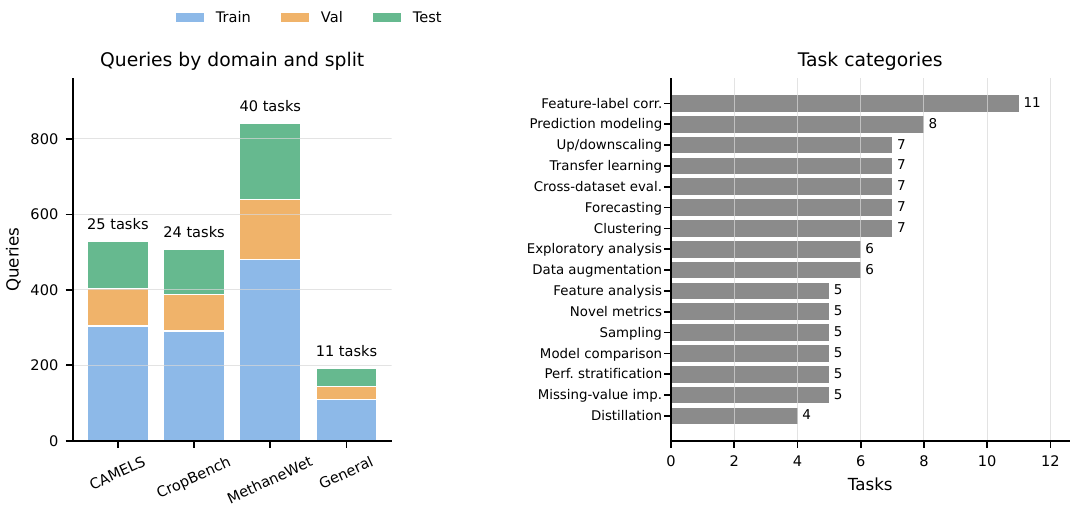}
\vspace{-20pt}
\caption{ST-Bench distribution. Left: query counts by domain and split, with task counts annotated above each domain. Right: the 16 task categories used to construct the benchmark.}
\label{fig:dataset-composition}
\vspace{-15pt}
\end{figure}

\begingroup\setlength{\abovedisplayskip}{3pt plus 1pt minus 1pt}\setlength{\belowdisplayskip}{3pt plus 1pt minus 1pt}\setlength{\abovedisplayshortskip}{2pt plus 1pt minus 1pt}\setlength{\belowdisplayshortskip}{2pt plus 1pt minus 1pt}

\section{Evaluation Framework}
\label{app:eval-framework}

This appendix provides additional details for the evaluation framework in Section~\ref{sec:eval}, including training protocols, metric definitions, the training-time reward function, and task annotation axes used for diagnostic analysis.

The core question we wish to answer thorough the evaluation is that whether MAS generation systems outperform single-agent baselines on real scientific problems, and if so by how much and at what cost. Answering it requires an evaluation framework that solves three problems specific to our setting. First, candidate systems produce open-ended numerical outputs rather than executable test results, so the test-time grader must reward real computation rather than reward defensive defaults. Second, MAS generation methods rely on optimizers that consume a scalar reward signal per candidate workflow, but scientific data analysis does not naturally provide such a signal. and scientific data analysis does not natively supply one.Third, a single composite score is not sufficient to reveal whether a workflow succeeds because it producesby achieving better metrics, because it coverscovering more queries, or only at a cost that erasesincurring costs that undermine its operational value. We design our evaluation framework around these three problems: a test-time grader with a realistic-output filter, a training-time verifier that delivers a continuous per-task reward signal, and a suite of MAS-specific evaluation metrics that diagnose performance along orthogonal dimensions.

\subsection{Training Protocols}
\label{app:training-protocols}

MAS generation methods differ in how they train and produce agentic workflows. Task-level methods optimize one MAS instance per narrow task using a training set of similar queries. Query-level methods maintain broader archives or controllers and produce one workflow per query. To compare both categories fairly, we evaluate every method under two protocols. In the per-task protocol, the method runs its inner-loop optimization separately on each of the 100 tasks, using only that task's train and validation queries, and produces 100 task-specific MAS instances. In the per-category protocol, the method optimizes on the union of train and validation queries from all tasks in a category, producing 16 category-level MAS instances. Both protocols evaluate on the same 492 held-out test queries from Section~\ref{sec:dataset-construction}.

\subsection{Evaluation Metrics}
\label{app:eval-metrics}

A single number per system tells who ranked higher but not why, or at what cost. Our evaluation reports one absolute performance score per system together with six comparative metrics relative to the single-agent baseline. Throughout, $M$ denotes a generated MAS configuration, $SA$ denotes the single-agent baseline, $T$ is the set of tasks, and $\mathrm{score}_{X,t}$ is the per-task absolute score of system $X$ on task $t$.

\paragraph{Absolute performance score.}
For each (task, query, metric) result, we check four conditions: parsability of the reported metrics, name match against a defined task metric, finite numeric validity, and satisfaction of the metric's inequality target if the metric needs a threshold, such as a minimum number of clusters that defines a successful clustering task outcome. These collect into a per-task success rate and a per-metric pass rate. To prevent defensive defaults from bypassing inequality grading, we apply a realistic filter where a query has realistic output if its reported metrics include at least one strictly positive numeric value and no $-1$ sentinel. The per-task absolute performance score is
{\small
\[
\mathrm{score}_t
=
\mathrm{success\_rate}_t
\cdot
\overline{\mathrm{pass\_rate}}_t ,
\]
}
where $\overline{\mathrm{pass\_rate}}_t$ is the mean per-metric pass rate over the task's realistic test queries (zero if the task has no realistic queries). The system's headline absolute score is the sum over all 100 tasks, with maximum 100. The score penalizes systems that fail entirely on a task more harshly than systems that succeed sparsely across many tasks, and ignores defensive zeros entirely.

\paragraph{Win rate and mean score gain.}
The fraction of tasks on which the MAS configuration improves over the single-agent baseline, paired with the average absolute improvement, is defined as
{\small
\[
\mathrm{win\_rate}(M)
=
\frac{1}{|T|}
\sum_{t \in T}
\mathbb{1}\bigl[
\mathrm{score}_{M,t} > \mathrm{score}_{SA,t}
\bigr],
\qquad
\Delta_{\mathrm{perf}}(M)
=
\frac{1}{|T|}
\sum_{t \in T}
\bigl(
\mathrm{score}_{M,t}
-
\mathrm{score}_{SA,t}
\bigr).
\]
}
Read together, the two metrics describe the shape of the improvement rather than only its sign. A high win rate paired with a small $\Delta_{\mathrm{perf}}$ identifies a system that improves many tasks by little; the opposite pairing identifies one that lifts a few tasks substantially. Either pattern can be desirable depending on the deployment context.

\paragraph{Cost ratio.}
The ratio of MAS inference cost to single-agent cost, measured along both wall-clock time and token consumption where the latter is available, is defined as
{\small
\[
\rho_{\mathrm{time}}(M)
=
\frac{\bar{c}^{\mathrm{time}}_{M}}
     {\bar{c}^{\mathrm{time}}_{SA}},
\qquad
\rho_{\mathrm{token}}(M)
=
\frac{\bar{c}^{\mathrm{token}}_{M}}
     {\bar{c}^{\mathrm{token}}_{SA}},
\]
}
where $\bar{c}^{\cdot}_X$ is the mean per-query cost of system $X$. We report $\rho_{\mathrm{time}}$ for every configuration because wall-clock time is logged uniformly across systems. Token accounting is consistently available for the single-agent baseline but not for every MAS adapter in the current release; we therefore report $\rho_{\mathrm{token}}$ where it is available and treat it as a partially measured axis.

\paragraph{Efficiency.}
The mean score gain per unit of additional inference cost is defined as
{\small
\[
\eta(M)
=
\frac{
\Delta_{\mathrm{perf}}(M)
}{
\max\bigl(\rho_{\mathrm{time}}(M) - 1,\, 0.1\bigr)
}.
\]
}
The denominator is clipped at $0.1$ to avoid division by zero on configurations whose cost ratio approaches one. Efficiency answers a question that single-agent benchmarks rarely have to ask: per unit of additional compute over the cheapest baseline, how much improvement does the MAS deliver? The same definition applies with $\rho_{\mathrm{token}}$ when token accounting is available.

\paragraph{Generalizability.}
The mean per-task minus per-category composite-score gap is defined as
{\small
\[
\mathrm{gen}(M)
=
\frac{1}{|T|}
\sum_{t \in T}
\bigl(
\mathrm{score}_{M^{\mathrm{task}},t}
-
\mathrm{score}_{M^{\mathrm{cat}},t}
\bigr),
\]
}
where $M^{\mathrm{task}}$ and $M^{\mathrm{cat}}$ are the per-task and per-category variants of the same method. A near-zero or negative value indicates that the broader category-level workflow transfers to sibling tasks at least as well as a task-specific workflow, while a strongly positive value indicates the method gains substantially from narrow specialization.

\paragraph{Scalability across backbones.}
This axis evaluates whether the architectural advantages of a generated MAS persist when the underlying LLM is replaced. To operationalize this, we select representative MAS configurations and evaluate their relative gains over single-agent baselines across multiple foundational models. Rather than an exhaustive sweep of all permutations, this representative sampling validates backbone scalability as a core, measurable dimension within the ST-Bench framework, establishing a standardized blueprint for future system audits.

\paragraph{Parallelism.}
Whether a generated MAS decomposes its work into subtasks that can execute concurrently. Wall-clock inference time relative to total accumulated agent time provides a proxy: a configuration whose wall-clock cost is substantially below its summed agent cost is exploiting parallelism, whereas a configuration whose wall-clock cost approximately equals its summed agent cost is sequential. The current release reports the wall-clock proxy; a full audit of internal subtask parallelization is left to future work.

These metrics are designed to be read together. A configuration with high win rate, high $\Delta_{\mathrm{perf}}$, low $\rho$, and high $\eta$ is unambiguously preferred. A configuration that scores high on $\Delta_{\mathrm{perf}}$ but with $\rho$ in the multiples is preferred only when absolute improvement matters more than its cost. The intra- and inter-system comparisons of \S\ref{sec:experiments} use this metric suite to support both leaderboard and diagnostic claims.

Because the 100 tasks span heterogeneous scientific domains and data-driven objectives, an aggregate score alone cannot reveal which kinds of tasks benefit most from MAS specialization, which is especially important for domain experts using MAS generation systems. We further annotate each task with difficulty and four orthogonal tag axes---capability dimension, pipeline stage, cognitive skill, and MAS challenge---so that results can be sliced by task structure. Full label definitions, counts, and examples are provided in Appendix~\ref{app:task-tags}.

\subsection{Training-time Reward Function}
\label{app:reward-function}

MAS generation methods drive their optimization with a scalar reward per candidate workflow. Coding benchmarks supply this reward through unit tests; scientific data analysis does not, and the absolute performance score of \S\ref{app:eval-metrics} is unsuitable as a training reward because all-zero and $-1$ defaults clear several inequality targets trivially and the optimizer would converge on them. We therefore define a continuous reference-aware reward function used in place of the absolute score during training.

For each defined metric of a query, the reward is the weighted sum
{\small
\[
s
=
0.10 \cdot s_{\mathrm{sanity}}
+
0.40 \cdot s_{\mathrm{target}}
+
0.50 \cdot s_{\mathrm{reference}},
\]
}
where $s_{\mathrm{sanity}}$ is $1$ for a finite numeric value, $s_{\mathrm{target}}$ is $1$ if the value satisfies the metric's inequality target, and $s_{\mathrm{reference}}$ is the proportion of the cached single-agent reference value the candidate reaches, clipped to $[0, 1]$ and polarity-flipped for lower-is-better metrics. A single override removes the defensive-zero attractor: a candidate value of zero against a non-zero reference returns score zero for that metric, regardless of the other components. For approximately seven percent of queries where the single-agent reference is missing, we drop $s_{\mathrm{reference}}$ and renormalize the remaining weights to $0.40$ and $0.60$. The per-query reward is the average of $s$ across the query's metrics.

The reward function is wired in evaluator-side: it replaces the per-query scoring hook that each method ordinarily calls during training, while leaving the prompt the candidate sees unchanged. This mirrors how each method consumes a binary oracle on coding benchmarks, and ensures any improvement we observe is attributable to better candidate selection rather than to leaking reference information into the prompt.

\subsection{Per-task Analytical Labels for Evaluation Cuts}
\label{app:task-tags}
 
The system metrics of \S\ref{sec:eval-axes} measure properties of the system; this subsection complements them with axes that slice the task pool itself. Together they form a 2D analysis grid: which property of the system are we measuring, on which subset of tasks. The labels we describe here are not part of the raw task definition (\S\ref{sec:dataset}); they are evaluation tooling that we add for analysis.
 
\paragraph{Difficulty.}
Each task carries a difficulty label set during annotation: 17 Easy, 49 Medium, and 34 Hard. Difficulty reflects the analytical complexity of the source publication. Easy tasks have a single self-contained metric and one canonical method. Hard tasks combine multiple metrics that depend on each other or require iterative refinement of intermediate artifacts. The difficulty mix is deliberately weighted toward Medium and Hard because MAS generation methods are unlikely to add value on Easy tasks where a single agent already succeeds; we want headline numbers dominated by tasks where MAS specialization is more likely to matter.
 
\paragraph{Four-axis groupings for MAS-targeted analysis.}
Each task additionally carries a four-axis tag set designed to expose orthogonal MAS-evaluation cuts.
 
\textbf{Capability dimension} (Modeling 31, Discovery 28, Optimization 23, Robustness 18) records the kind of capability the task exercises: building predictive models, discovering structure in data, solving an explicit objective under constraints, or maintaining performance under noise and shift.
 
\textbf{Pipeline stage} (Modeling 33, Data Analysis 25, Evaluation 23, Data Preparation 19) records where in the data-science pipeline the task sits, so we can ask which pipeline stages most benefit from MAS role decomposition. This grouping lines up with how MAS designers describe the internal role decomposition of a generated MAS (a ``data-prep agent'' or a ``model-builder agent''), and lets us probe directly whether handing a particular stage to a dedicated agent inside the MAS pays off on real scientific data.
 
\textbf{Cognitive skill} (Predictive 39, Diagnostic 33, Prescriptive 16, Descriptive 12) records the analytical style the task asks for, following the standard analytics taxonomy: what is happening (Descriptive), why is it happening (Diagnostic), what will happen (Predictive), what should we do (Prescriptive). The skill axis is informative for MAS evaluation because Diagnostic and Prescriptive tasks typically require coordinating multiple sub-analyses, which is exactly what a generated MAS can in principle do better than a single agent.
 
\textbf{MAS challenge} (Synthesis 35, Parallel Search 23, Long-horizon Planning 23, Error Recovery 14, Decomposition 5) records the kind of MAS-specific competency the task most stresses. Tasks tagged ``Decomposition'' require breaking the problem into independent subproblems before any agent can act. Tasks tagged ``Parallel Search'' admit many candidate approaches that benefit from concurrent exploration. Tasks tagged ``Long-horizon Planning'' require maintaining state and intermediate results across many steps. Tasks tagged ``Synthesis'' require aggregating partial results from multiple sub-analyses into a coherent final answer. Tasks tagged ``Error Recovery'' deliberately surface failure modes that the system must catch and re-try. We use the MAS-challenge axis in \S\ref{sec:experiments} to ask whether the methods that score well overall do so because they handle a specific type of MAS challenge particularly well, or because they perform uniformly across all five.

\section{Experiments and Results}
\begin{table}[t]
\centering
\caption{Per-task training budgets.}
\vspace{-5pt}
\label{tab:budgets}
\begin{tabular}{lll}
\toprule
Method & Inner loop & Per-task budget \\
\midrule
MetaAgent  & one-shot FSM generation         & 1 round (no iterative selection) \\
AutoAgents & per-query Explorer              & 1 round, 3 test queries cap \\
GPTSwarm   & edge-realization optimization   & 3 realizations $\times$ 2 val queries \\
AFlow      & MCTS over operator graphs       & 2 rounds, 2 val sample, 1 validation round \\
W4S        & iterative meta-agent proposals  & 3 iterations $\times$ 2 val queries \\
\bottomrule
\end{tabular}
\end{table}

\subsection{Coverage-Quality Decomposition Details}
\label{app:coverage-quality}

The absolute performance score is the product of two factors that we can examine separately: the realistic output rate, defined as the fraction of test queries on which the system produces a non-defensive metrics dictionary, and the mean pass rate conditional on realistic output, defined as the share of inequality targets cleared on those queries. Figure~\ref{fig:cost-coverage}b plots each MAS configuration on these two axes.

Two patterns dominate the figure. First, conditional pass rates cluster within a relatively narrow band, typically between 50\% and 58\%, while realistic output rates vary much more widely across methods and protocols. Second, the configurations that achieve the highest absolute scores in \S\ref{sec:exp-headline} do so by raising the realistic output rate rather than the conditional quality. W4S per-task reaches the highest realistic rate at 38.2\%; GPTSwarm and AFlow configurations cluster at intermediate realistic rates; the AutoAgents per-category configuration sits at the lowest realistic rate, consistent with its query-cap-induced coverage shortfall. The narrow band of conditional pass rates indicates that, once a system reaches the point of producing realistic metrics, the share of inequality targets it clears is comparable across methods; the differentiator is whether the system reaches that point at all.

This decomposition refines the finding that MAS specialization improves the absolute score primarily through coverage rather than through higher metric quality. It also identifies where future MAS design effort is most likely to translate directly into absolute-score gains. Improvements that raise the realistic output rate---data-loading robustness, exception-handling discipline that does not collapse to defensive defaults, and end-to-end execution over staged files---move the system along the dominant axis of variation. Improvements aimed at fine-grained metric quality run into the narrow conditional band and yield smaller absolute-score gains.

\subsection{Stratified Results: Domain, Tags, and Difficulty}
\label{app:exp-stratified}

The result in Section~\ref{sec:exp-headline} averages over heterogeneous scientific tasks. We therefore stratify the same composite test-split score by three metadata axes: scientific domain, task tags, and declared difficulty. For each group, the GPT-5 score is the mean per-task composite score in that group. The MAS bars and table entries select the best verifier-trained MAS configuration within the group.

\begin{figure}[t]
\centering
\includegraphics[width=\linewidth]{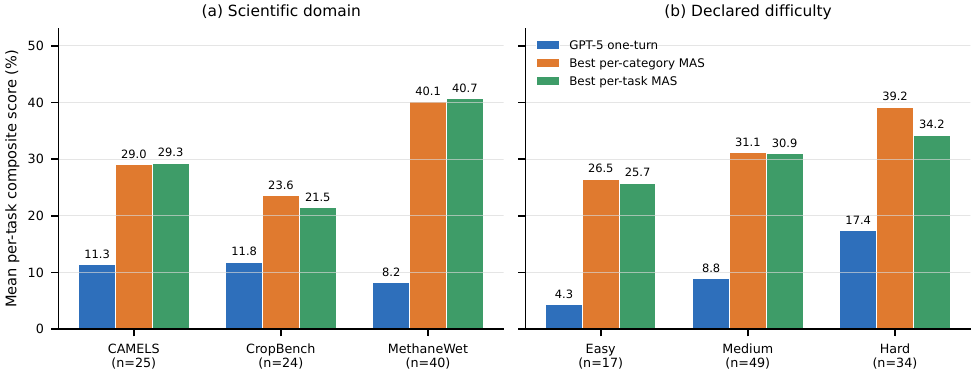}
\vspace{-15pt}
\caption{Task-subset cuts by scientific domain and declared difficulty. Each MAS bar is the best verifier-trained configuration of that training granularity within the group.}
\label{fig:stratified-cuts}
\vspace{-15pt}
\end{figure}

\paragraph{Domain.}
Figure~\ref{fig:stratified-cuts}a focuses on the three scientific domains in ST-Bench. MAS improves all three, but not uniformly. The largest gain appears on MethaneWet: GPT-5 scores only 8.2 on average, while GPTSwarm reaches 40.1 under per-category training and 40.7 under per-task training. CAMELS also benefits strongly, with GPTSwarm per-category and AFlow per-task both near 29. CropBench is the weakest MAS domain: GPTSwarm per-category reaches 23.6 and W4S per-task reaches 21.5, still roughly doubling the GPT-5 one-turn baseline but with a smaller margin than the other scientific domains. This suggests that MAS generation is especially useful when the task family requires repeated data loading, modeling, and metric computation over similar structured workflows, as in MethaneWet and CAMELS.

\paragraph{Tag system.}
Table~\ref{tab:tag-stratified} stratifies tasks by the benchmark's four tag axes: capability dimension, pipeline stage, cognitive skill, and MAS challenge. The strongest gains occur in groups that should naturally reward structured multi-agent decomposition. Data preparation gains 35.2 pp, robustness gains 34.7 pp, decomposition gains 34.0 pp, and parallel search gains 33.0 pp. These are precisely the settings where a generated workflow can separate file inspection, transformation, modeling, and validation steps. The smallest gains are still positive, but more modest: error recovery gains 17.1 pp, pipeline-stage modeling gains 19.0 pp, and synthesis gains 20.3 pp. This pattern is useful for benchmark interpretation: MAS generation is not just "better on average"; it is most valuable on tags that expose coordination, search, and data-handling structure.

\begin{table*}[t]
\centering
\small
\vspace{-5pt}
\caption{Best MAS configuration by task-tag group. Scores are mean per-task composite score percentages. Gain is best MAS minus GPT-5 one-turn in percentage points; win rate is the fraction of tasks in the group where the selected MAS configuration beats GPT-5.}
\resizebox{\textwidth}{!}{%
\begin{tabular}{llrrrrr}
\toprule
Tag axis & Tag value & Tasks & GPT-5 & Best MAS & Gain & Win rate \\
\midrule
Capability & Discovery & 28 & 11.2 & GPTSwarm cat 38.6 & 27.4 & 57.1\% \\
Capability & Modeling & 31 & 9.9 & W4S cat 32.4 & 22.5 & 54.8\% \\
Capability & Optimization & 23 & 18.0 & GPTSwarm task 41.1 & 23.1 & 56.5\% \\
Capability & Robustness & 18 & 3.5 & W4S task 38.2 & 34.7 & 61.1\% \\
Cognitive & Descriptive & 12 & 10.0 & GPTSwarm cat 41.7 & 31.7 & 58.3\% \\
Cognitive & Diagnostic & 33 & 10.2 & W4S task 35.5 & 25.3 & 57.6\% \\
Cognitive & Predictive & 39 & 8.7 & W4S cat 38.5 & 29.8 & 59.0\% \\
Cognitive & Prescriptive & 16 & 19.0 & W4S task 44.0 & 25.1 & 68.8\% \\
MAS challenge & Decomposition & 5 & 6.0 & AFlow task 40.0 & 34.0 & 40.0\% \\
MAS challenge & Error recovery & 14 & 8.0 & MetaAgent cat 25.1 & 17.1 & 57.1\% \\
MAS challenge & Long-horizon planning & 23 & 16.2 & W4S cat 47.2 & 31.0 & 60.9\% \\
MAS challenge & Parallel search & 23 & 6.1 & GPTSwarm task 39.1 & 33.0 & 60.9\% \\
MAS challenge & Synthesis & 35 & 12.7 & W4S task 33.0 & 20.3 & 57.1\% \\
Pipeline & Data analysis & 25 & 11.7 & GPTSwarm cat 42.4 & 30.7 & 60.0\% \\
Pipeline & Data preparation & 19 & 7.0 & W4S task 42.2 & 35.2 & 73.7\% \\
Pipeline & Evaluation & 23 & 7.1 & W4S cat 35.9 & 28.8 & 60.9\% \\
Pipeline & Modeling & 33 & 15.4 & GPTSwarm task 34.4 & 19.0 & 51.5\% \\
\bottomrule
\end{tabular}}
\label{tab:tag-stratified}
\vspace{-15pt}
\end{table*}

\paragraph{Difficulty.}
Figure~\ref{fig:stratified-cuts}b shows that MAS gains persist across Easy, Medium, and Hard tasks. The declared difficulty labels are not monotonic with one-turn baseline performance: Easy tasks have the lowest GPT-5 one-turn score at 4.3, while Hard tasks have the highest at 17.4. This is likely because the labels describe scientific task complexity, whereas the one-turn baseline is dominated by execution brittleness and whether the first attempt reaches a realistic metric computation. MAS improves all levels by roughly 22 pp under the best configuration. The best per-category MAS is slightly stronger than the best per-task MAS at every difficulty level in this aggregate, suggesting that broader category-level workflows often capture reusable structure even when individual tasks are labeled difficult.


\end{document}